\documentclass{article}
\usepackage{spconf}
\usepackage[utf8]{inputenc}
\usepackage{tikz}
\usepackage{pgfplots}
\usepackage{graphicx}
\usepackage{todonotes}
\usepackage{adjustbox}
\usepackage{multirow}
\usepackage{multicol}

\pgfplotsset{compat=1.16}
\usetikzlibrary{pgfplots.colorbrewer,}
\pgfplotsset{
    cycle list/.define={my marks}{
        every mark/.append style={solid,fill=\pgfkeysvalueof{/pgfplots/mark list fill}},mark=*\\
        every mark/.append style={solid,fill=\pgfkeysvalueof{/pgfplots/mark list fill}},mark=square*\\
        every mark/.append style={solid,fill=\pgfkeysvalueof{/pgfplots/mark list fill}},mark=triangle*\\
        every mark/.append style={solid,fill=\pgfkeysvalueof{/pgfplots/mark list fill}},mark=diamond*\\
    },
}
\pgfplotsset{
  errorBars/.style={
    error bars/error bar style={very thick},
    error bars/error mark options={very thick,solid,mark size=3pt,rotate=90},
    error bars/y dir=both,
    error bars/y explicit,
  }
}
\pgfplotsset{
  log ticks with fixed point,
}
\def\addlegendimage{\csname pgfplots@addlegendimage\endcsname}
\usepgfplotslibrary{external}
\title{Noise Masking Attacks and Defenses for Pretrained Speech Models}

\name{Matthew Jagielski \qquad Om Thakkar \qquad Lun Wang}
  
\address{Google}

\begin{document}

\maketitle

\begin{abstract}
Speech models are often trained on sensitive data in order to improve model performance, leading to potential privacy leakage. Our work considers noise masking attacks, introduced by Amid et al. \cite{amid2022extracting}, which attack automatic speech recognition (ASR) models by requesting a transcript of an utterance which is partially replaced with noise. They show that when a record has been seen at training time, the model will transcribe the noisy record with its memorized sensitive transcript. 
In our work, we extend these attacks beyond ASR models, to attack pretrained speech encoders. Our method fine-tunes the encoder to produce an ASR model, and then performs noise masking on this model, which we find recovers private information from the pretraining data, despite the model never having seen transcripts at pretraining time! We show how to improve the precision of these attacks and investigate a number of countermeasures to our attacks.
\end{abstract}


\section{Introduction}
A common paradigm for building more performant and robust speech models is by building foundation models through large scale pretraining. Such models build a strong general understanding of speech, and can later be fine-tuned for specific speech applications~\cite{bestrq,zhang2023google}. Pretraining generally requires large datasets, where it is important to reach for diverse, and potentially sensitive, data sources. However, this carries a risk: machine learning models in general have been shown to be susceptible to privacy attacks which leak information about their training data~\cite{shokri2017membership, carlini2021extracting}.

One compelling form of privacy leakage which is specific to speech models is the noise masking attack, introduced by Amid et al. \cite{amid2022extracting} to attack automated speech recognition (ASR) models. 
Rather than considering privacy attacks which may result in relatively limited harm, such as membership inference~\cite{shokri2017membership}, or considering extraction for unnatural ``canary'' examples~\cite{huang2022detecting}, a noise masking attack directly demonstrates leakage on real data records. To perform such an attack, an adversary who knows some nonsensitive subset of an utterance can fill the sensitive subset with noise, such as cafe noise or even silence, and input it to the ASR model. For utterances contained in the training data of the ASR model, the model will sometimes transcribe the noise with the exact text seen during training, revealing the sensitive portion of the utterance. 
For example, if the adversary queries with an audio of the form ``Mister $\textsc{noise}$ has Lyme disease'', the model can produce a transcription replacing the noise with the real subject of this training utterance.

As proposed by Amid et al. \cite{amid2022extracting}, noise masking can only be applied to ASR models, on data where the model has trained on both audio and a transcript. In our work, we extend this to the modern setting of pretraining on large, potentially sensitive datasets. In particular, we consider models trained with self-supervised learning, on data which consists only of audio, and so, the model \emph{never learns from sensitive text}. To do so, we design an attack which first carefully fine-tunes a pretrained encoder, using different data, to build an ASR model. Our key finding is that this approach leads to successful noise masking, paralleling privacy attacks which have been shown on the pretraining data for image models \cite{guo2023ssl, abascal2023tmi}. We also experiment with mitigations. In summary, our contributions are:
\begin{enumerate}
    \item We extend the noise masking attacks of Amid et al. \cite{amid2022extracting} to modern large scale pretraining, and show how to increase the precision of any noise masking attack.
    \item We design several mitigations for our attacks, based on prior defenses for noise masking as well as mitigations for extraction attacks proposed for non-speech models.
\end{enumerate}

\section{Noise Masking on Pretrained Encoders}
\begin{figure}
    \centering
    \includegraphics[width=\linewidth]{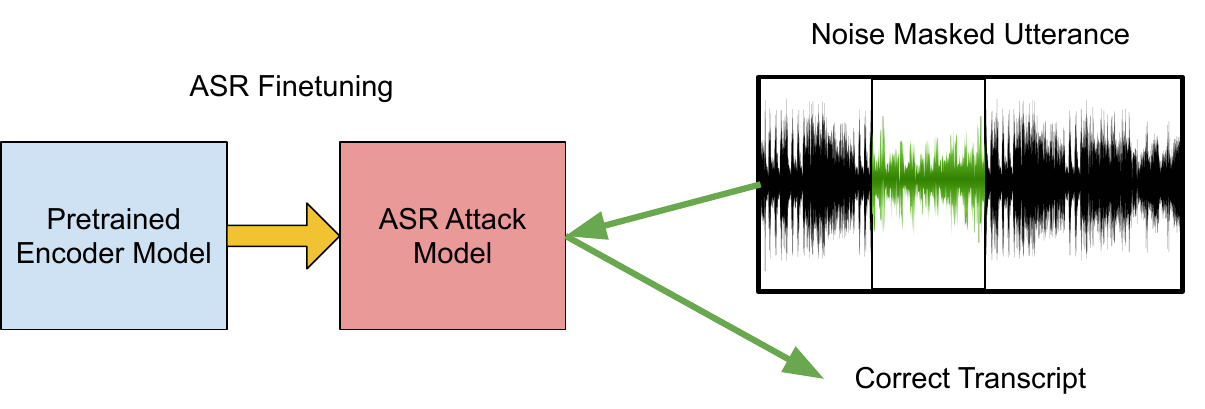}
    \caption{Our noise masking attack pipeline for recovering sensitive information from pretraining data. We take a pretrained encoder model, fine-tune it to produce an ASR attack model, and run noise masking on that attack model.}
    \label{fig:nm_fig}
\end{figure}
\label{sec:attack}
We consider an adversary with access to an audio encoder model $E$ which has been trained on a pretraining dataset consisting of audio examples $X_{PT}$, which is sampled from some underlying population of pretraining data $P_{PT}$. The adversary's goal is to learn sensitive information about examples in $X_{PT}$. However, this encoder takes as input an audio example and ``encodes'' it into some high dimensional, difficult-to-interpret latent space, so understanding the concrete privacy risk of these examples is difficult with only access to this encoder. In order to relax this constraint, we also allow the adversary to collect a dataset of transcribed audio for ASR fine tuning $(X_{FT}, Y_{FT})$, which does not overlap with the sensitive audio in $X_{PT}$. We also carry over several assumptions from the existing noise masking work:
\begin{enumerate}
    \item The adversary has domain knowledge which allows them to identify target utterances, containing both nonsensitive and sensitive portions, where the former portion can be easily produced, but the latter is unknown (e.g., being able to produce ``Mister $\textsc{unknown}$ lives in Seoul'', but not knowing the sensitive name). 
    \item The adversary has a function $N$, which adds noise to an utterance, in place of the sensitive utterance portion.
\end{enumerate}

\textbf{Our Noise Masking Attack.}
Our approach begins by taking the encoder model $E$ and performing supervised fine tuning on it using $(X_{FT}, Y_{FT})$, producing an ASR model $M$. From here, we can perform noise masking directly on $M$ as proposed in Amid et al.~\cite{amid2022extracting}. One significant challenge in our work is that continued training, as done in fine tuning, has been shown to reduce privacy leakage on prior examples \cite{jagielski2022measuring}, which in our setting is the privacy leakage on $X_{PT}$.

\textbf{High Precision Noise Masking.}
While noise masking attacks have been traditionally designed to maximize overall attack success rate, the privacy literature has recently begun to allow adversaries the ability to ``abstain'' from predictions when they are not confident about the result of an attack \cite{leino2020stolen, carlini2022membership}. When the adversary abstains from predictions, attack success rate can be evaluated only on trials without abstention. This captures a setting where leakage is more harmful if it is more actionable (i.e., an adversary has higher confidence in the information).

We extend noise masking attacks to maximize precision as well, by allowing our adversary to abstain on low confidence attack results. We consider two different ways for an adversary to abstain from a prediction: transcript-based and agreement-based. In transcript-based abstention, the adversary filters out attack outputs which transcribe only the nonsensitive audio. In an agreement-based abstention, the adversary tries multiple types of noises, and only keeps predictions where multiple noise types provide the same transcript. We define how we measure precision in Section~\ref{ssec:exp-setup}.


\section{Experiments}
\label{sec:exp}
We now evaluate the effectiveness of our noise masking attacks. We first describe our experiment setup, and then we evaluate three research questions.

\subsection{Experiment Setup}
\label{ssec:exp-setup}
We use the LibriLight dataset \cite{LL20} for pretraining (i.e., $X_{PT}$), and the LibriSpeech (LS) dataset \cite{LS15} for fine tuning (i.e., $(X_{FT}, Y_{FT})$). We use these as they are standard datasets for evaluating pretraining and fine tuning. One important consideration for these datasets is their overlap - because LibriSpeech overlaps with LibriLight, if we pretrain a model on LibriLight and fine-tune on LibriSpeech, a noise masking attack on a record from LibriLight could rely on memorization from the a similar record in LibriSpeech, overestimating risk to the pretraining data. 
To overcome this confounder, we produce a filtered version of LibriSpeech (LS-NoName) without \emph{any} formatted names\footnote{We find formatted names by searching for the titles ``mister'', ``miss'', and ``misses'' in LS transcripts, and filter out those cases where the next word is a common word.}. Then, any successful noise masking attack extracting formatted names must come as a result of memorization of the name seen during pretraining.

All of our models use the BEST-RQ~\cite{bestrq} pretraining recipe on a 600M parameter encoder containing a stack of 24 Conformer~\cite{conf20} models, and are fine-tuned for ASR with CTC loss~\cite{graves2012connectionist}.
All training parameter values are set following the original BEST-RQ setup.
Unless otherwise specified, we run pretraining for 1 million steps.

To run our attacks, we use roughly 1000 utterances containing names from the LibriSpeech training set (i.e., not contained in LS-NoName), and all of the roughly 80 LibriSpeech test examples which contain names. 
Following the setup in \cite{amid2022extracting}, we consider noise masking with 5 sampled noises per utterance, from ‘car’, cabin noise, ‘cafe’ chatter with background
noise, ‘music’, ‘kitchen’ background noise, and ‘podcast’.  We also separately consider silence masking.

\textbf{Metrics.}
We consider multiple metrics, computed over 5 trials. We also sometimes report Word Error Rate (WER) after fine-tuning to measure encoder quality.
\begin{enumerate}
    \item ``Exact name'' accuracy - the fraction of utterances for which any attack (i.e., any of the 5 noise samples) successfully recovers the correct name. This reflects the risk to an average sensitive pretraining example.
    \item ``Exact name'' precision - the fraction of utterances for which any non-abstained attack successfully recovers the correct name. This reflects how confident an adversary can be in the name produced by an attack.
    \item ``Any name'' accuracy - the fraction of utterances for which any name is returned by noise masking. This means someone's private information has leaked, even if not the exact target.
    \item ``Any name'' precision - the fraction of utterances for which any non-abstained attack successfully recovers any name. Private information has leaked, and the adversary is confident that it is private.
\end{enumerate}

\subsection{Results}
We organize our results around answering three questions.
\textbf{RQ1: Is it possible to perform a noise masking attack on pretraining data?}
We run our attacks on two models, fine-tuned on LS and LS-NoName for 10,000 steps. We report the results of noise masking in Table~\ref{tab:rq1}. Noise masking attacks are possible on the pretraining data. When finetuning on both LS and LS-NoName, we find our attacks can correctly recover the exact name from roughly 1-2\% of training utterances, even without any overlap with the finetuning set (for LS-NoName), and leakage of any name is much higher.

Recall that there is an overlap between LibriSpeech and LibriLight, so results with LS finetuning can be the result of memorization of both pretraining and finetuning data, rather than just pretraining data in LS-NoName. However, even this result is somewhat positive, as attacks on end-to-end ASR training in Amid et al.~\cite{amid2022extracting} result in much higher success rates. For example, with silence masking, Amid et al.~\cite{amid2022extracting} recover correct names from 11.8\% of training utterances, while we recover correct names from only 2.7\% of training utterances (although our best attacks are with noise masking). This is likely because more epochs are required to train a good ASR model from scratch compared to a pretrained encoder; Amid et al.~\cite{amid2022extracting} train for 100,000 steps, 10 times longer than our models, which allows much more memorization.

\begin{table}[]
    \centering
    \begin{tabular}{cc|ccc}
        \multirow{2}{*}{Finetune} & \multirow{2}{*}{Attack} & \multicolumn{2}{c}{Train} & Test \\
        ~ & ~ & Exact & Any & Any\\
        \hline
        LS & Noise & 2.7/2.6 & 11.9/14.2 & 5.7/9.3 \\
        LS & Silence & 1.4/1.5 & 6.5/9.3 & 2.9/3.7 \\
        LS-NoName & Noise & 1.6/1.5 & 10.8/13.8 & 2.9/1.9 \\
        LS-NoName & Silence & 1.1/1.0 & 4.9/7.3 & 0/1.9\\
    \end{tabular}
    \caption{It is possible to run a noise masking attack to recover sensitive information seen during pretraining. Each cell contains reported values for the clean/other split from LibriSpeech. We report both exact name accuracy and any name accuracy, measured in percentages. Test Exact results are omitted as they are all 0: the model cannot exactly complete data it has never seen.}
    \label{tab:rq1}
\end{table}

\noindent
\textbf{RQ2: How precise can noise masking attacks be?}
For the attacks in RQ1, we also evaluate their precision using transcript-based and agreement-based filtering as described in Section 2, reporting the best precisions achieved by these approaches in Table~\ref{tab:rq2}. All of the best results use transcript-based filtering, although sometimes combining transcript and agreement-based filtering outperforms transcript-based filtering alone. Precision can be much higher than accuracy, as transcript and agreement filtering can reliably identify unlikely continuations. For example, silence masking attacks can reach over 12\% utterance precision, when these attacks had under 2\% accuracy without filtering.

\begin{table}[]
    \centering
    \begin{tabular}{cc|ccc}
        \multirow{2}{*}{Finetune} & \multirow{2}{*}{Attack} & \multicolumn{2}{c}{Train} & Test\\
        ~ & ~ & Exact & Any & Any \\
        \hline
        LS & Noise & 4.7/4.0 & 17.0/19.0 & 15.6/15.2 \\
        LS & Silence & 2.5/2.6 & 35.9/45.5 & 33.3/35.2 \\
        LS-NoName & Noise & 2.1/3.0 & 12.8/20.5 & 3.6/0 \\
        LS-NoName & Silence & 8.5/7.6 & 25.6/37.9 & 0/0 \\
    \end{tabular}
    \caption{Noise masking can be made more precise (often substantially) by abstaining from attack predictions where the adversary has low confidence. We report the best filtering strategy for each result. Transcript-based filtering is always effective, and combining it alignment-based filtering can sometimes improve precision further. Each cell contains precision, in percentages, for the clean/other split from LibriSpeech. Test Exact results are all 0, as in Table~\ref{tab:rq1}.}
    \label{tab:rq2}
\end{table}

\noindent
\textbf{RQ3: Are there any modeling/training choices which impact the risk of the attack to the pretraining data?}
We experiment with pretraining length and codebook dimension for pretraining. Intuitively, shorter pretraining results in less effective noise masking, due to less ability to memorize pretraining data --- pretraining for only 20k steps instead of 1M steps (50 times shorter) results in worse attacks (accuracy of 0.1/0.2), but also significantly higher WER (8.7 instead of 3.6). We also experiment with changing BestRQ parameters such as the codebook dimension and vocab size, which control the size of the random projections used for tokenization in BestRQ pretraining. In general, we find that changing these parameters from their defaults (codebook dimension of 16, and vocab size of 8192) has little impact on noise masking risk, except for when model performance is harmed. For a very small codebook dimension of 2, WER increases from 4.0 to 4.6, and reduces exact name accuracy to only 0.1\% under silence masking. For the remaining models, attacks perform comparably to the results reported in Table~\ref{tab:rq1}.

\section{Mitigations}
\label{sec:defense}

We consider a number of mitigations for our attacks, inspired by those proposed in prior work to prevent noise masking directly, and those that have been considered in the literature for defending against other types of extraction attacks.

\subsection{Our Mitigations}
\textbf{Data Sanitization.} Data curation has been proposed in prior work to mitigate extraction attacks~\cite{carlini2021extracting, lukas2023analyzing}. Intuitively, removing sensitive data from pretraining will prevent it from being memorized. However, such curation can be difficult, as curating before training requires anticipating all potential sensitivities in the dataset. Furthermore, detecting such sensitivity at scale is challenging, as it may require training an additional model, such as a sensitivity classifier, and running that model over all pretraining data. We propose here an ideal defense for our noise masking attacks, where the learner removes all names from LibriLight. To construct this dataset, we automatically transcribe all of LibriLight with an ASR model, and then filter out all utterances whose transcripts contain any formatted name, similar to how we created LS-NoName earlier.

\noindent
\textbf{Modified Pretraining.} Amid et al.~\cite{amid2022extracting} show that modifying ASR training by adding silences and noise (through MTR \cite{kim2017generation}) can reduce the model's propensity to complete noise masked utterances. Our learner cannot influence the ASR model's training, so we experiment instead with using MTR (with the same parameters as in \cite{amid2022extracting}) and silencing during pretraining. Both these approaches can encourage the pretrained model to predict noise/silence when given noisy/silent context, potentially reducing how frequently the fine-tuned model ``overconfidently'' makes predictions with a noise masking attack. For silence masking during pretraining, we consider adding 100ms and 500ms of silence, but only report results for 100ms silence due to the limited impact of this parameter.

\noindent
\textbf{Data Deduplication.}
Prior work has shown that deduplicating training data can mitigate extraction risk in language models~\cite{lee2021deduplicating, kandpal2022deduplicating} and diffusion models~\cite{carlini2023extracting}. However, these works consider extraction of entire records or large portions of records, while our focus is on extracting small subrecords. To deduplicate sensitive data, such as names, we choose to deduplicate our pretraining dataset by removing highly duplicated $k$-grams in the transcripts of utterances. Our deduplication approach, which we call DD-$(k, p)$ has two parameters: $k$, the length of $k$-grams to use for deduplication, and $p$, the fraction of the dataset to remove. That is, we remove the most highly duplicated $k$-grams until a $p$ fraction of the dataset has been removed. Note that names may not be the most duplicated $k$-grams, so even large values of $p$ may remove few names. We vary $k$ from 3 and 5, and set $p$ to remove no more than 10\% of the training set. For $3$-grams, we find that removing 35\% of the dataset is required to remove even 1\% of $k$-grams containing names, so we set $p=0.35$ for $k=3$.

\subsection{Results}
We run pretraining with each mitigation, and apply our noise masking attack to each resulting encoder after fine tuning on LS-NoName. We report the attack accuracies in Table~\ref{tab:mitigations}. We do not report silence masking results due to their qualitatively similar results to noise masking. We measure encoder quality with the ASR model's WER on Librispeech test-other partition, but among mitigations we find limited WER difference.

Our most successful mitigation is data sanitization, with near-zero exact name accuracy. Despite removing all names, data sanitization does not perfectly prevent the attack, likely due to some imperfection in our name filtering or inaccuracies in our transcription which we use to filter.

We also observe small exact name attack success rates when modifying pretraining with MTR and when combined with silencing. This parallels the effectiveness of a comparable combined strategy from \cite{amid2022extracting}. Interestingly, silencing alone is not effective, even against silencing attacks. This could be the result of ``catastrophic forgetting'' \cite{mccloskey1989catastrophic} during fine tuning of the silencing behavior seen during pretraining.

Deduplication has limited effectiveness at mitigating our attacks, with attack success rates as high or even higher than unmitigated results. This poor performance is likely the result of many names remaining in the pretraining data, due to not being duplicated enough to be removed. For example, by removing highly duplicated $3$-grams to remove $35\%$ of the pretraining data, we deduplicate only $5\%$ of $3$-grams with names. Diverse pretraining data could contain sensitive $k$-grams of varying lengths, so deduplication is unlikely to be a panacea for dealing with privacy risks in speech data.

No mitigation significantly reduces the ``any name'' accuracy. This could be the result of the model repeating names found in non-formatted contexts. We also remark that these accuracies are still lower than those found for the best mitigation on end-to-end ASR training from \cite{amid2022extracting}.

\begin{table}[]
    \centering
    \begin{tabular}{c|c|ccc}
        \multirow{2}{*}{Mitigation} & \multirow{2}{*}{WER} & \multicolumn{2}{c}{Train} & Test \\
        ~ & ~ & Exact & Any & Any \\
        \hline
         None (Baseline) & 4.0 & 1.6/1.5 & 10.8/13.8 & 2.9/1.9 \\
        Sanitization & 4.4 & 0/0.3 & 10.0/12.0 & 8.6/3.7 \\
        Silence & 4.0 & 2.9/1.3 & 13.5/14.5 & 5.7/7.4 \\
        MTR & 4.6 & 0.4/0.4 & 12.7/13.6 & 8.6/7.4 \\
        Silence+MTR & 4.0 & 0.1/0.3 & 13.8/14.8 & 5.7/1.9 \\
        DD-$(3, 0.35)$ & 4.0 & 0.6/1.1 & 10.4/14.1 & 5.7/7.4 \\
        DD-$(4, 0.05)$ & 3.8 & 1.4/0.5 & 12.3/14.1 & 5.7/5.6 \\
        DD-$(4, 0.1)$ & 4.0 & 2.3/1.3 & 13.2/15.2 & 8.6/7.4 \\
        DD-$(5, 0.05)$ & 4.0 & 1.6/1.3 & 9.8/13.5 & 14.3/1.9 \\
    \end{tabular}
    \caption{Noise masking can be mitigated. Our best mitigations are silence masking + MTR and data sanitization, while data deduplication and silence masking are not very effective.}
    \label{tab:mitigations}
\end{table}

\section{Conclusion}
Our work finds that privacy leakage is possible on sensitive data seen only during pretraining time. Our attacks are less successful than those on end-to-end ASR models trained on sensitive data as in prior work, but due to the continued risk, we propose data curation and training techniques to limit the risks of privacy attacks. Stronger attacks than those we devise may be possible using more sophisticated noise generation or attack model training, so we also recommend continued research into privacy-preserving training techniques.

\newpage
\bibliographystyle{IEEEtran}
\bibliography{refs.bib}

\begin{thebibliography}{10}
\providecommand{\url}[1]{#1}
\csname url@samestyle\endcsname
\providecommand{\newblock}{\relax}
\providecommand{\bibinfo}[2]{#2}
\providecommand{\BIBentrySTDinterwordspacing}{\spaceskip=0pt\relax}
\providecommand{\BIBentryALTinterwordstretchfactor}{4}
\providecommand{\BIBentryALTinterwordspacing}{\spaceskip=\fontdimen2\font plus
\BIBentryALTinterwordstretchfactor\fontdimen3\font minus
  \fontdimen4\font\relax}
\providecommand{\BIBforeignlanguage}[2]{{%
\expandafter\ifx\csname l@#1\endcsname\relax
\typeout{** WARNING: IEEEtran.bst: No hyphenation pattern has been}%
\typeout{** loaded for the language `#1'. Using the pattern for}%
\typeout{** the default language instead.}%
\else
\language=\csname l@#1\endcsname
\fi
#2}}
\providecommand{\BIBdecl}{\relax}
\BIBdecl

\bibitem{amid2022extracting}
E.~Amid, O.~D. Thakkar, A.~Narayanan, R.~Mathews, and F.~Beaufays, ``Extracting
  targeted training data from {ASR} models, and how to mitigate it,'' in
  \emph{Interspeech 2022, 23rd Annual Conference of the International Speech
  Communication Association}, 2022.

\bibitem{bestrq}
C.~Chiu, J.~Qin, Y.~Zhang, J.~Yu, and Y.~Wu, ``Self-supervised learning with
  random-projection quantizer for speech recognition,'' in \emph{International
  Conference on Machine Learning, {ICML} 2022}, 2022.

\bibitem{zhang2023google}
Y.~Zhang, W.~Han, J.~Qin, Y.~Wang, A.~Bapna, Z.~Chen, N.~Chen, B.~Li,
  V.~Axelrod, G.~Wang \emph{et~al.}, ``Google usm: Scaling automatic speech
  recognition beyond 100 languages,'' \emph{arXiv preprint arXiv:2303.01037},
  2023.

\bibitem{shokri2017membership}
R.~Shokri, M.~Stronati, C.~Song, and V.~Shmatikov, ``Membership inference
  attacks against machine learning models,'' in \emph{2017 IEEE symposium on
  security and privacy (SP)}.\hskip 1em plus 0.5em minus 0.4em\relax IEEE,
  2017, pp. 3--18.

\bibitem{carlini2021extracting}
N.~Carlini, F.~Tramer, E.~Wallace, M.~Jagielski, A.~Herbert-Voss, K.~Lee,
  A.~Roberts, T.~Brown, D.~Song, U.~Erlingsson \emph{et~al.}, ``Extracting
  training data from large language models,'' in \emph{30th USENIX Security
  Symposium (USENIX Security 21)}, 2021.

\bibitem{huang2022detecting}
W.~R. Huang, S.~Chien, O.~D. Thakkar, and R.~Mathews, ``Detecting unintended
  memorization in language-model-fused {ASR},'' in \emph{Interspeech 2022, 23rd
  Annual Conference of the International Speech Communication
  Association}.\hskip 1em plus 0.5em minus 0.4em\relax {ISCA}, 2022, pp.
  2808--2812.

\bibitem{guo2023ssl}
C.~Guo, F.~Bordes, P.~Vincent, and K.~Chaudhuri, ``Do ssl models have
  d$\backslash$'ej$\backslash$a vu? a case of unintended memorization in
  self-supervised learning,'' \emph{arXiv preprint arXiv:2304.13850}, 2023.

\bibitem{abascal2023tmi}
J.~Abascal, S.~Wu, A.~Oprea, and J.~Ullman, ``Tmi! finetuned models leak
  private information from their pretraining data,'' \emph{arXiv preprint
  arXiv:2306.01181}, 2023.

\bibitem{jagielski2022measuring}
M.~Jagielski, O.~Thakkar, F.~Tramer, D.~Ippolito, K.~Lee, N.~Carlini,
  E.~Wallace, S.~Song, A.~G. Thakurta, N.~Papernot \emph{et~al.}, ``Measuring
  forgetting of memorized training examples,'' in \emph{The Eleventh
  International Conference on Learning Representations}, 2022.

\bibitem{leino2020stolen}
K.~Leino and M.~Fredrikson, ``Stolen memories: Leveraging model memorization
  for calibrated $\{$White-Box$\}$ membership inference,'' in \emph{29th USENIX
  security symposium (USENIX Security 20)}, 2020, pp. 1605--1622.

\bibitem{carlini2022membership}
N.~Carlini, S.~Chien, M.~Nasr, S.~Song, A.~Terzis, and F.~Tramer, ``Membership
  inference attacks from first principles,'' in \emph{2022 IEEE Symposium on
  Security and Privacy (SP)}.\hskip 1em plus 0.5em minus 0.4em\relax IEEE,
  2022, pp. 1897--1914.

\bibitem{LL20}
J.~Kahn, M.~Rivi{\`{e}}re, W.~Zheng, E.~Kharitonov, Q.~Xu, P.~Mazar{\'{e}},
  J.~Karadayi, V.~Liptchinsky, R.~Collobert, C.~Fuegen, T.~Likhomanenko,
  G.~Synnaeve, A.~Joulin, A.~Mohamed, and E.~Dupoux, ``Libri-light: {A}
  benchmark for {ASR} with limited or no supervision,'' in \emph{2020 {IEEE}
  International Conference on Acoustics, Speech and Signal Processing, {ICASSP}
  2020}.

\bibitem{LS15}
V.~Panayotov, G.~Chen, D.~Povey, and S.~Khudanpur, ``Librispeech: An {ASR}
  corpus based on public domain audio books,'' in \emph{2015 {IEEE}
  International Conference on Acoustics, Speech and Signal Processing, {ICASSP}
  2015}.\hskip 1em plus 0.5em minus 0.4em\relax {IEEE}, 2015, pp. 5206--5210.

\bibitem{conf20}
A.~Gulati, J.~Qin, C.~Chiu, N.~Parmar, Y.~Zhang, J.~Yu, W.~Han, S.~Wang,
  Z.~Zhang, Y.~Wu, and R.~Pang, ``Conformer: Convolution-augmented transformer
  for speech recognition,'' in \emph{Interspeech 2020, 21st Annual Conference
  of the International Speech Communication Association}.\hskip 1em plus 0.5em
  minus 0.4em\relax {ISCA}, 2020, pp. 5036--5040.

\bibitem{graves2012connectionist}
A.~Graves and A.~Graves, ``Connectionist temporal classification,''
  \emph{Supervised sequence labelling with recurrent neural networks}, pp.
  61--93, 2012.

\bibitem{lukas2023analyzing}
N.~Lukas, A.~Salem, R.~Sim, S.~Tople, L.~Wutschitz, and
  S.~Zanella-B{\'e}guelin, ``Analyzing leakage of personally identifiable
  information in language models,'' \emph{arXiv preprint arXiv:2302.00539},
  2023.

\bibitem{kim2017generation}
C.~Kim, A.~Misra, K.~K. Chin, T.~Hughes, A.~Narayanan, T.~N. Sainath, and
  M.~Bacchiani, ``Generation of large-scale simulated utterances in virtual
  rooms to train deep-neural networks for far-field speech recognition in
  google home,'' in \emph{Interspeech 2017, 18th Annual Conference of the
  International Speech Communication Association}.\hskip 1em plus 0.5em minus
  0.4em\relax {ISCA}, 2017, pp. 379--383.

\bibitem{lee2021deduplicating}
K.~Lee, D.~Ippolito, A.~Nystrom, C.~Zhang, D.~Eck, C.~Callison-Burch, and
  N.~Carlini, ``Deduplicating training data makes language models better,''
  \emph{arXiv preprint arXiv:2107.06499}, 2021.

\bibitem{kandpal2022deduplicating}
N.~Kandpal, E.~Wallace, and C.~Raffel, ``Deduplicating training data mitigates
  privacy risks in language models,'' in \emph{International Conference on
  Machine Learning}.\hskip 1em plus 0.5em minus 0.4em\relax PMLR, 2022, pp.
  10\,697--10\,707.

\bibitem{carlini2023extracting}
N.~Carlini, J.~Hayes, M.~Nasr, M.~Jagielski, V.~Sehwag, F.~Tramer, B.~Balle,
  D.~Ippolito, and E.~Wallace, ``Extracting training data from diffusion
  models,'' in \emph{32nd USENIX Security Symposium (USENIX Security 23)},
  2023.

\bibitem{mccloskey1989catastrophic}
M.~McCloskey and N.~J. Cohen, ``Catastrophic interference in connectionist
  networks: The sequential learning problem,'' in \emph{Psychology of learning
  and motivation}.\hskip 1em plus 0.5em minus 0.4em\relax Elsevier, 1989,
  vol.~24, pp. 109--165.

\end{thebibliography}

\end{document}